%% file: aaai21.tex
\def\year{2021}\relax
\documentclass[letterpaper]{article} 
\usepackage{aaai21}  
\usepackage{times}  
\usepackage{helvet} 
\usepackage{courier}  
\usepackage[hyphens]{url}  
\usepackage{graphicx} 
\usepackage{natbib}  
\usepackage{caption} 
\usepackage{booktabs}       
\usepackage{nicefrac}       
\usepackage{float}
\input{math_commands.tex}

\usepackage{tikz}
\usetikzlibrary{shapes,arrows,fit,chains,matrix}
\usetikzlibrary{positioning, shapes.geometric}
\tikzstyle{node} = [circle, minimum size = 18mm, thick, draw =black!80]
\tikzstyle{nodeinter} = [rectangle, minimum size = 16mm, thick, draw =black!80, fill=gray!30]
\tikzstyle{nodeinterwhite} = [rectangle, minimum size = 16mm, thick, draw =black!80]
\tikzstyle{nodeobserved} = [circle, minimum size = 18mm, thick, draw =black!80, fill=gray!30]
\tikzstyle{box} = [rectangle, draw =black!0]
\tikzstyle{arrow} = [thick,->,>=stealth,line width=0.6mm]
\tikzstyle{arrow2} = [dashed,->,>=stealth]

\title{Selecting Data Augmentation for Simulating Interventions}
\author {

        Maximilian Ilse,\textsuperscript{\rm 1}
        Jakub M. Tomczak, \textsuperscript{\rm 2}
        Patrick  Forr\'e \textsuperscript{\rm 1} \\
}
\affiliations {
    \textsuperscript{\rm 1} Amsterdam Machine Learning Lab,
University of Amsterdam \\
    \textsuperscript{\rm 2} Computational Intelligence Group,
Vrije Universiteit Amsterdam \\
    m.ilse@uva.nl
}

\begin{document}

\maketitle

\begin{abstract}
Machine learning models trained with purely observational data and the principle of empirical risk minimization \citep{vapnik_principles_1992} can fail to generalize to unseen domains. In this paper, we focus on the case where the problem arises through spurious correlation between the observed domains and the actual task labels. We find that many domain generalization methods do not explicitly take this spurious correlation into account. Instead, especially in more application-oriented research areas like medical imaging or robotics, data augmentation techniques that are based on heuristics are used to learn domain invariant features. To bridge the gap between theory and practice, we develop a causal perspective on the problem of domain generalization. We argue that causal concepts can be used to explain the success of data augmentation by describing how they can weaken the spurious correlation between the observed domains and the task labels. We demonstrate that data augmentation can serve as a tool for simulating interventional data. We use these theoretical insights to derive a simple algorithm that is able to select data augmentation techniques that will lead to better domain generalization.
\end{abstract}

\noindent

\section{Introduction}
Despite recent advancements in machine learning fueled by deep learning, studies like \citet{azulay_why_2019} have shown that deep learning methods may not generalize to inputs from outside of their training distribution.
In safety-critical fields like medical imaging, robotics and, self-driving cars, however, it is essential that machine learning models are robust to changes in the environment. Without the ability to generalize, machine learning models cannot be safely deployed in the real world.

In the field of domain generalization, one tries to find a representation that generalizes across different environments, called \textit{domains}, each with a different shift of the input. This problem is especially challenging when changes in the domain are spuriously associated with changes in the actual task labels. This can, for instance, happen when the data gathering process is biased. An example is given by \citet{arjovsky_invariant_2019}: If we consider a dataset of images of cows and camels in their natural habitat, then there is a strong correlation between the type of animal and the landscape in the image, e.g., a camel standing in a desert. If we now train a machine learning model to predict the animal in a given image, the model is prone to exploit the spurious correlation between the type of animal and the type of landscape. As a result, the model can fail to recognize a camel standing in a green pasture or a cow standing in a desert.

In recent years, a large corpus of methods designed to learn representations that will generalize across domains has been formulated. While the proposed methods are able to achieve good results on a variety of domain generalization benchmarks, the majority of them lack a theoretical foundation. In the worst-case scenario, these methods enforce the wrong type of invariance, as proven in the Appendix. Interestingly, we find that especially in more applied fields, like medical imaging and robotics, researchers have found a practical way of dealing with the spurious correlation between domains and the actual task. Data augmentation in combination with Empirical Risk Minimization (ERM) \citep{vapnik_principles_1992} is used to enforce invariance of the machine learning model with respect to changes in the domain. Hereby, prior knowledge is used to guide the selection of appropriate data augmentation. In the Appendix, we give a detailed summary of two successful applications of data augmentation in the context of domain generalization.

However, the success of data augmentation is often described in vague terms like 'artificially expanding labeled training datasets' \citep{li_automating_2020} and 'reduce overfitting' \citep{krizhevsky_imagenet_2012}. In this paper, we present a causal perspective on data augmentation in the context of domain generalization and contribute to the field in the following manner:
\begin{itemize}

    \item First, we introduce the concept of \textit{intervention-augmentation equivariance} that formalizes the relationship between data augmentation and interventions on features caused by the domain. We show that if intervention-augmentation equivariance holds we can use data augmentation to successfully simulate interventions using only observational data.

    \item Second, we derive a simple algorithm that is able to select data augmentation techniques from a given list of transformations. We compare our approach to a variety of domain generalization methods on three domain generalization benchmarks. We demonstrate that we are able to consistently outperform all other methods.
\end{itemize}

\section{Method}
\label{sec:method}
\subsection{Domain generalization}
\label{sec:dom_gen_def}
We first formalize the problem of domain generalization following the notations used in \citet{muandet_domain_2013}.
We assume that during training we have access to samples $\mathcal{S}$ from $N$ different domains, where $\mathcal{S} = \{S^{d=i}\}^N_{i=1}$. From each domain $n_i$ samples $S^{d=i} = \{(x^{d=i}_k, y^{d=i}_k)\}^{n_i}_{k=1}$ are included in the training set. The training data is represented as tuples of the form $(x,y,d)$ sampled from the observational distribution $ p(x,y,d)$. The goal of domain generalization is to develop machine learning methods that generalize well to unseen domains. In order to test the ability of a machine learning model to generalize, we use samples $S^{d=N+1}$ from a previously unseen test domain $d=N+1$.

In this paper, we are interested in the general case where the observed domains $d$ and targets $y$ are spuriously correlated in the training dataset, i.e., where we might have $p(y|d=i)\neq p(y|d=j), i,j \in \{1, \dots, N\}.$ Since the correlation between $d$ and $y$ is assumed to be spurious, it does not necessarily hold for the test domain $d=N+1$.

\subsection{Domain generalization and data augmentation from a causal perspective}
For readers unfamiliar with the concepts of causality, a brief introduction of the causal concepts that are used throughout this paper can be found in the Appendix. For an in-depth introduction please see \citet{pearl_causal_2009} or \citet{Peters2017}.

First, we introduce a Structural Causal Model (SCM) in order to describe what we believe in many cases reflects the underlying causal structure of domain generalization problems.
\begin{figure}[h]
\begin{minipage}{.265\textwidth}
\centering
    \resizebox{4.2cm}{4.7cm}{
    \input{tikz_plots/confounder_proxies.tex}
    }
\end{minipage}
\begin{minipage}{.2\textwidth}
\begin{align}
    d &:= f_D(c) \nonumber\\
    y &:= f_Y(c) \nonumber\\
    h_d &:= f_{H_d}(d) \nonumber\\
    h_y &:= f_{H_y}(y) \nonumber\\
    x &:= f_X(h_d, h_y),
    \label{eq:causal_process}
\end{align}
\end{minipage}
    \caption{DAG and SCM with a hidden confounder.}
    \label{fig:gen_w_conf}
\end{figure}
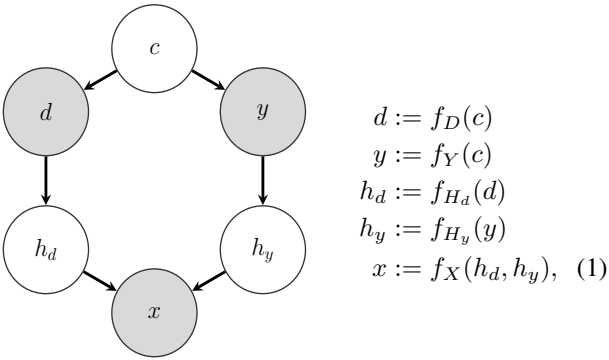
The SCM is shown in Figure \ref{fig:gen_w_conf} (right) where $c$ is a hidden confounder (and a exogenous variable), $d$ the domain, $y$ the target,  $h_d$ high-level features, e.g., color and orientation, caused by $d$, $h_y$ high level-features, e.g., shape and texture, caused by $y$, and $x$ the input. We omit including noise variables for clarity. The corresponding Directed Acyclic Graph (DAG) is shown in Figure \ref{fig:gen_w_conf} (left), where a  grey node means the variable is observed and a white node corresponds to a latent (unobserved) variable. The presented DAG is similar to the ones constructed in \citet{subbaswamy_development_2019} and \citet{castro_causality_2019}. In Figure \ref{fig:gen_w_conf}, the node $c$ is a hidden confounder. The hidden confounder $c$ opens up a backdoor path (a non-causal path) $d \xleftarrow{} c \xrightarrow{}
y$ \citep{pearl_causal_2009}. This path allows $d$ to enter $y$ trough the back door.

As a result, the domain $d$ and the target $y$ are in general no longer independent, $p(y,d)$ $\neq$ $p(y)p(d)$. Since the high-level features, $h_d$ are children of $d$, they are spuriously correlated with $y$ as well, i.e., $h_d$ becomes predictive of $y$. We now assume that we train a machine learning model using ERM \citep{vapnik_principles_1992} and observational data generated from the DAG in Figure \ref{fig:gen_w_conf}. The task is to predict $y$ from $x$, which itself is anti-causal. Since $d$ and $y$ are correlated, it is likely that the machine learning model will rely on all high-level features $h_d$ and $h_y$ to predict $y$. Furthermore, we assume that the correlation of $d$ and $y$ is spurious. Therefore, it will not hold in general and will break under intervention. A machine learning model relying on high-level features $h_d$ that are caused by $d$ is thus likely to generalize poorly to unseen domains. Returning to our introductory example of classifying animals in images, the hidden confounder can be used to model the fact that there is a common cause for the type of animal and the landscape in an image. For example, the confounder could be the country in which a particular image was taken, e.g., in Switzerland we are more likely to see a cow standing in a green pasture than a camel or a desert.

\subsection{Simulating interventions}
\label{sec:sim_inter}
\begin{figure*}
    \begin{minipage}{.32\textwidth}
    \centering
    \resizebox{4.5cm}{4.9cm}{
    \input{tikz_plots/intervention_on_d.tex}
    }
    \end{minipage}
    \begin{minipage}{.32\textwidth}
    \centering
    \resizebox{4.5cm}{4.9cm}{
    \input{tikz_plots/intervention_on_hd.tex}
    }
    \end{minipage}
    \begin{minipage}{.32\textwidth}
    \centering
    \resizebox{4.5cm}{6.7cm}{
    \input{tikz_plots/data_augmentation.tex}
    }
    \end{minipage}
    \caption{Left: DAG with hidden confounder after intervention on $d$. Center: DAG with hidden confounder after intervention on $h_d$. Interventional nodes are squared. Right: DAG with hidden confounder plus data augmentation. Note that in the latter case we do not have to intervene on the system that generates the data. Data augmentation should be designed in a way such that the augmented data simulates data from the center or left DAG.}
    \label{fig:gen_w_conf_inter}
\end{figure*}
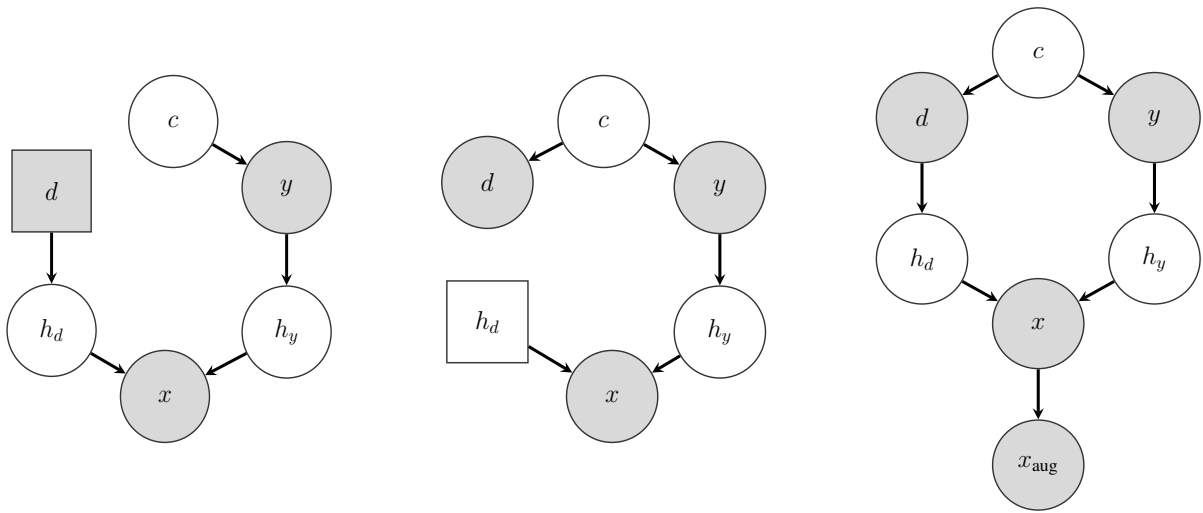

One possible approach to deal with the spurious correlations between $d$ and $y$ is to perform an intervention on $d$. Such an intervention would render $d$ and $y$ independent, i.e., $p(y| \doit(d))= p(y)$. In Figure \ref{fig:gen_w_conf_inter} (left), we see the same DAG as in Figure \ref{fig:gen_w_conf} but after we intervened on $d$. We find that in Figure \ref{fig:gen_w_conf_inter} (left) there is no more arrow connecting the hidden confounder $c$ and the domain $d$. The backdoor path $d \xleftarrow{} c \xrightarrow{} y$ has vanished. In the examples of animals and landscapes, to intervene on the domain $d$ (the landscape), we would have to physically move a cow to a desert. It becomes apparent that the interventions have to happen in the real world and are not operations on the already gathered observational data. In the majority of domain generalization problems, it will not be feasible to collect new data with specific interventions. 

In Figure \ref{fig:gen_w_conf_inter} (center) we present a second way of addressing the problem of correlated variables $d$ and $y$. In theory one could perform an intervention on all high-level features $h_d$, i.e., $\doit(h_d)$, since $d$ affects $x$ only indirectly via $h_d$, in our example $h_d$ could represent the colors and textures of the landscapes. Again, an intervention like this would need to happen during the data collection process in the real world, e.g., by moving sand to a pasture.

However, we argue that in certain cases we can simulate data from the interventional distribution $p(x,y|\doit(h_d))$ using data augmentation in combination with observational data. For example, we could randomly perturb the colors in the animal images. This type of augmentation simulates a noise intervention on $h_d$, i.e., $\doit(h_d=\xi)$, where $\xi$ is sampled from a noise distribution $N_\xi$ \citep{peters_causal_2016}. By augmenting only high-level features $h_d$ that are caused by $d$ we guarantee that the target $y$ and features $h_y$ are unchanged. After data augmentation the pairs $(x_\text{aug},y)$ should closely resemble samples from the interventional distribution $p(x,y|\doit(h_d))$. In Figure \ref{fig:gen_w_conf_inter} (right) we see that we only require observational data from the DAG without any interventions. While each augmented sample $x_\text{aug}$ individually can be seen as a counterfactual, we argue that we effectively marginalize over the counterfactual distribution by generating a multitude of augmented samples $x_\text{aug}$ from each $x$. We argue that for correctly designed data augmentation we cannot distinguish the data generated by any of the three models in Figure \ref{fig:gen_w_conf_inter}.
In theory, we could intervene on $h_d$ by setting $h_d$ to a fixed value, instead of performing a noise intervention. However, in order to simulate data from such an interventional distribution using data augmentation, we would require $h_d$ to be observed, which we argue is generally not the case. In the Appendix, we describe that there exist data augmentation methods that try to infer $h_d$ for each sample $x$ before setting $h_d$ to a fixed value for all samples, yet these augmentations seem to perform worse than randomly sampled augmentations.\\

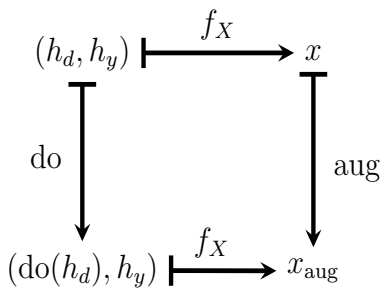
\begin{figure}[h]
    \centering
    \resizebox{5.5cm}{4cm}{
    \input{tikz_plots/commutative.tex}
    }
    \caption{Intervention-augmentation equivariance expressed in a commutative diagram. }
    \label{fig:commutative}
\end{figure}

If we want to design data augmentation $x_\aug=\aug(x)$, as a transformation $\aug(\cdot)$ applied to observed data $x$, such that it simulates an intervention on the high-level features $h_d$ caused by $d$, one needs to make assumption about the causal data generating process. Formally, we require that augmenting the data $x$ to $x_\aug=\aug(x)$ commutes with an intervention $\doit(h_d)$ prior to the data generation. We call this \emph{intervention-augmentation equivariance}. In more formal detail, assume that we have the causal process from Equation \ref{eq:causal_process}: $x := f_X(h_d, h_y)$.
Then augmenting $x$ via $\aug(\cdot)$ does:
\begin{align}
    x_\text{aug} &= \aug(x) \nonumber\\
    &= \aug(f_X(h_d, h_y)).
\end{align}
We then say that the causal process $f_X: \mathcal{H}_d \times \mathcal{H}_y \mapsto \mathcal{X}$, is \emph{intervention-augmentation equivariant} if for every considered stochastic data augmentation transformation $\aug(\cdot)$ on $x \in \mathcal{X}$ we have a corresponding noise intervention $\doit(\cdot)$ on $h_d \in \mathcal{H}_d$ such that:
\begin{align}
    \aug(f_X(h_d, h_y)) = f_X(\doit(h_d), h_y).
    \label{eq:data_aug}
\end{align}
The intervention-augmentation equivariance is expressed as a commutative diagram in Figure \ref{fig:commutative}. We argue that by making strong assumptions about the true causal process we need to first identify the high-level features $h_d$ caused by $d$. Second, we have to design data augmentation $\aug(x)$ that commutes with a corresponding intervention $\doit(h_d)$ under the causal process $f_X(h_d, h_y)$.
A special case of intervention-augmentation equivariance occurs in the classical case of an $G$-equivariant map $f_X$, where $G$ can be any (semi-)group. For this to hold, we need $G$ to act on the spaces $\mathcal{H}_y$, $\mathcal{H}_d$, $\mathcal{X}$, and we need to make sure that $G$ acts trivially on $\mathcal{H}_y$. So any element $g \in G$ can transform elements $x \in \mathcal{X}$ into $g\cdot x \in \mathcal{X}$, which we will interpret as data augmentation, as demonstrated in the Experiment section. The elements $g \in G$ also transform  $h_d \in \mathcal{H}_d$ into $g\cdot h_d \in \mathcal{H}_d$, which we consider as a special type of intervention. Furthermore, $h_y \in \mathcal{H}_y$ are assumed to be kept fixed $g\cdot h_y = h_y$ for all $g \in G$. So we put:
\begin{align}
    \doit(h_d) &:= g\cdot h_d,\\
    \aug(x) &:= g\cdot x,
\end{align}
where we assume that the elements $g \in G$ are randomly sampled from some distribution $p(g)$ on $G$.
In this setting, any \emph{$G$-equivariant map} $f_X$ is then automatically also intervention-augmentation equivariant, as can be seen from:
\begin{align}
\aug(x) &= g\cdot f_X(h_d, h_y)\\
&=f_X(g\cdot h_d, g\cdot h_y)\\
&= f_X(\doit(h_d), h_y),
                 \label{eq:group}
\end{align}
a linear example of intervention-augmentation equivariance can be found in the Appendix. 

In general, we find that the majority of frequently used data augmentations can be expressed as simple group actions. For example, randomly rotating the input image $x$ can be understood as randomly sampling and applying elements $g$ from the two-dimensional rotation group $SO(2)$ on the two dimensional pixel grid. Randomly changing the hue of an image $x$ corresponds to randomly sampling and applying elements $g$ from the two-dimensional rotation group $SO(2)$, since hue can be represented as an angle in color space. Applying random permutations to the color channels of an image $x$ is equivalent to randomly sampling and applying elements $g$ from permutation group $S_3$, in the case of three separate color channels.

\subsection{Selecting data augmentations for domain generalization}
\label{sec:select_da}
In Figure \ref{fig:gen_w_conf_inter} (center), we see that if we successfully simulate an intervention on $h_d$ using data augmentation the arrow from $d$ to $h_d$ vanishes.
Based on this theoretical insight, we propose an algorithm that is able to select data augmentation techniques that will improve domain generalization, instead of manually choosing them. In the following we will refer to the algorithm as Select Data Augmentation (SDA).  Similar to \citet{cubuk_autoaugment_2019}, we start with a list of data augmentation techniques including: 'brightness', 'contrast', 'saturation', 'hue', 'rotation', 'translate', 'scale', 'shear', 'vertical flip', and 'horizontal flip'. Since these transformations do not influence each other, they can be tested separately. The hyperparameter for each augmentation can be found in the Appendix. The proposed SDA algorithm consists of three steps:
\begin{enumerate}
    \item We divide all samples from the training domains into a training and validation set.
    \item We train a classifier to predict the domain $d$ from input $x$. During training, we apply the first data augmentation in our list to the samples of the training set. We save the domain accuracy on the validation set after training. We repeat this step with all data augmentations in the list.
    \item We select the data augmentation with the lowest domain accuracy averaged over five seeds. If multiple data augmentations lie within the standard error of the selected one they are selected as well, i.e., there is no statistically significant difference between the augmentations.
\end{enumerate}
Intuitively, SDA will select data augmentation techniques that destroy information about $d$ in $x$. From a causal point of view, this is equivalent to weaken the arrow from $d$ to $h_d$. In the Appendix, we perform an ablation study showing that SDA also reliably selects the most suitable data augmentation if the list contains the same augmentation with different hyperparameters.

There is one caveat though. Throughout this entire section, we assume that we are successfully augmenting all high-level features $h_d$ caused by $d$. In a real-world application, we usually have no means to validate this assumption, i.e., we might only augment a subset of $h_d$. Furthermore, we might even augment high-level features $h_y$ that are caused by the target node $y$. Nonetheless, we argue there are cases where we still obtain better generalization performance than a machine learning model trained without data augmentation. This may happen in cases where weakening the spurious confounding influence of $h_d$ on $y$ recovers more of the anti-causal signal for $y$ than the data augmentation on the features $h_y$ destroys. We evaluate this hypothesis empirically in the Experiment section.

\section{Related work}
\label{sec:related_work}
\subsection{Learning symmetries from data} In the previous section, we argue that choosing the right symmetry group for data augmentation relies on prior knowledge, e.g., preselecting the list of transformations to test. While this is a clear practical limitation of our approach, to the best of our knowledge there exist no approaches that are able to learn symmetries from purely observational data. Contemporary approaches like Lagrangian neural networks \citep{cranmer_lagrangian_2020}, graph neural networks \citep{kipf_semi-supervised_2017}, and group equivariant neural networks \citep{cohen_group_2016} are enforcing apriori chosen symmetries instead of learning them.

\subsection{Understanding data augmentation}
Recently, \citet{gontijo-lopes_affinity_2020} develop two measures:  affinity
and diversity. The measures are used to quantify the effectiveness of existing data augmentation methods. They find that augmentations that have high affinity and diversity scores lead to better generalization performance. While affinity
and diversity rely on the iid assumption, we provide an alternative for non-iid datasets. \citet{lyle_benefits_2020} investigate how data augmentation can be used to incorporate invariance into machine learning models. They show that while data augmentation can lead to tighter PAC-Bayes bounds, data augmentation is not guaranteed to lead to invariance. In Equation, \ref{eq:data_aug} we formalize under which condition (namely intervention-augmentation equivariance) data augmentation will lead to invariance.

\subsection{Advanced data augmentation techniques}
\citet{zhang_mixup_2018} introduced a method called mixup that constructs new training examples by linearly interpolating between two existing examples ($x_i$, $y_i$) and ($x_j$, $y_j$). In \citet{gowal_achieving_2019} and \citet{perez_effectiveness_2017} a Generative Adversarial Network (GAN) is used to perform so-called 'adversarial mixing'. The GAN is able to generate new training examples that belong to the same class $y$ but have different styles. Furthermore, \citet{perez_effectiveness_2017} propose a novel method called 'neural augmentation' where they train the first part of their model to generate an augmented image from two training examples with the same class $y$.
\subsection{Causality}
In \citet{peters_causal_2016} a method for Invariant Causal Prediction (ICP) is developed. It is built on the assumption that causal features are stable given different experimental settings. Given the complete set of causal features, the conditional distribution of the target variable $y$ must remain the same under interventions, e.g., change of the domain. Whereas, predictions made by a machine learning model relying on non-causal features are in general not stable under interventions. Recently, \citet{arjovsky_invariant_2019} proposed a framework called Invariant Risk Minimization (IRM), that shares the same goal as ICP. In IRM a soft penalty in combination with an ERM term is used to balance the invariance and predictive power of the learned machine learning model. In contrast to ICP, IRM can be used for tasks on unstructured data, e.g., images. However, while both methods (ICP and IRM) try to learn features that are parents of $y$, we argue that for the majority of domain generalization problems the task of predicting $y$ from $x$ is anti-causal. Therefore we are interested in augmenting only features caused by $d$, i.e., the descendants of $d$, assuming that the remaining features are caused by $y$.
In \citet{arjovsky_invariant_2019}, they argue that there exists a discrepancy between the true label (part of the true causal mechanism) that caused $x$ and the annotation produced by human labelers. Learning this 'labeler function' will lead to a good generalization performance, even though it might rely on patterns that are anti-causal or non-causal. In this situation, the IRM objective becomes ineffective.

\citet{heinze-deml_conditional_2019} introduced the Conditional variance Regularization
(CoRe). CoRe uses grouped observations (e.g., training samples with the same class $y$ but different styles) to learn invariant representations. Samples are grouped by an additional ID variable, which is different from the label $y$. We find that in most cases it is difficult to obtain an additional ID variable, e.g., none of the datasets in the Experiment section features such a variable. If no such ID variable exists, CoRe can use pairs of original images and augmented images to learn invariant representations.

While we are focusing on the DAG in Figure \ref{fig:gen_w_conf}, \citet{bareinboim_causal_2016} and \citet{mooij_joint_2019} have developed general graphical representations for relating data generating processes across domains. If the confounder $c$ was observed methods that find stable feature sets such as those in \citet{rojas-carulla_invariant_nodate} and \citet{magliacane_domain_2018}, could be used. Furthermore, \citet{subbaswamy_preventing_2019} shows that instead of intervening in some cases, it is possible to fit an interventional distribution from observational data. However, imaging data poses a challenge that existing causal-based methods are not equipped to deal with thus motivating the use of data augmentation.

\section{Experiments}
\label{sec:experiments}
We evaluate the performance of data augmentation in combination with Empirical Risk Minimization (ERM) \citep{vapnik_principles_1992} on four datasets. While the first is a synthetic dataset, the other three are domain generalization benchmark image datasets (rotated MNIST, colored MNIST, and PACS) where the domain $d$ and target $y$ are confounded. For the benchmark image datasets, we first use SDA to select the best data augmentation techniques. The results for this first step can be found in Table \ref{tab:domain_acc_results} in the Appendix. Afterwards, we apply the selected data augmentations and train the respective model using ERM. Finally, we perform an ablation study where we apply all data augmentations to all three image datasets and instead of the selected ones. 

Code to replicate all experiments can be found under \url{https://github.com/AMLab-Amsterdam/DataAugmentationInterventions}.

\subsection{Synthetic data}
\label{sec:toy_ex}
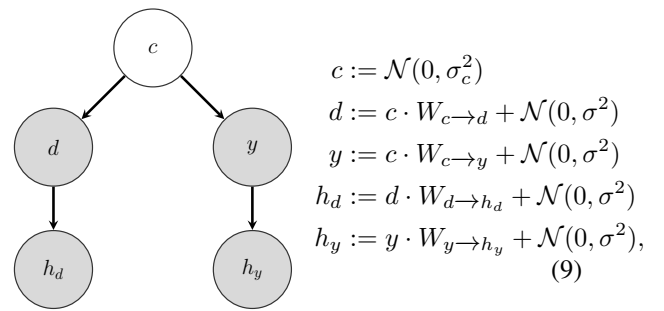
\begin{figure}[h]
\begin{minipage}{.22\textwidth}
    \resizebox{3.8cm}{4.0cm}{%
    \input{tikz_plots/experiment_dag.tex}
    }
\end{minipage}
\begin{minipage}{.2\textwidth}
\centering
\begin{align}
    c &:= \mathcal{N}(0, \sigma_c^2) \nonumber\\
    d &:= c \cdot W_{c\xrightarrow{}d} + \mathcal{N}(0, \sigma^2) \nonumber\\
    y &:= c \cdot W_{c\xrightarrow{}y} + \mathcal{N}(0, \sigma^2) \nonumber\\
    h_d &:= d \cdot W_{d\xrightarrow{}h_d} + \mathcal{N}(0, \sigma^2) \nonumber\\
    h_y &:= y \cdot W_{y\xrightarrow{}h_y} + \mathcal{N}(0, \sigma^2),
\end{align}
\end{minipage}
   \caption{DAG and linear Gaussian SCM for synthetic data.}
    \label{fig:toy_dag_and_scm}
\end{figure}

For the first experiment we simulate data from the linear Gaussian SCM in Figure \ref{fig:toy_dag_and_scm} (right), where the corresponding DAG can be seen in Figure \ref{fig:toy_dag_and_scm} (left).

\begin{figure}[h]
    \centering
    \includegraphics[scale=0.58]{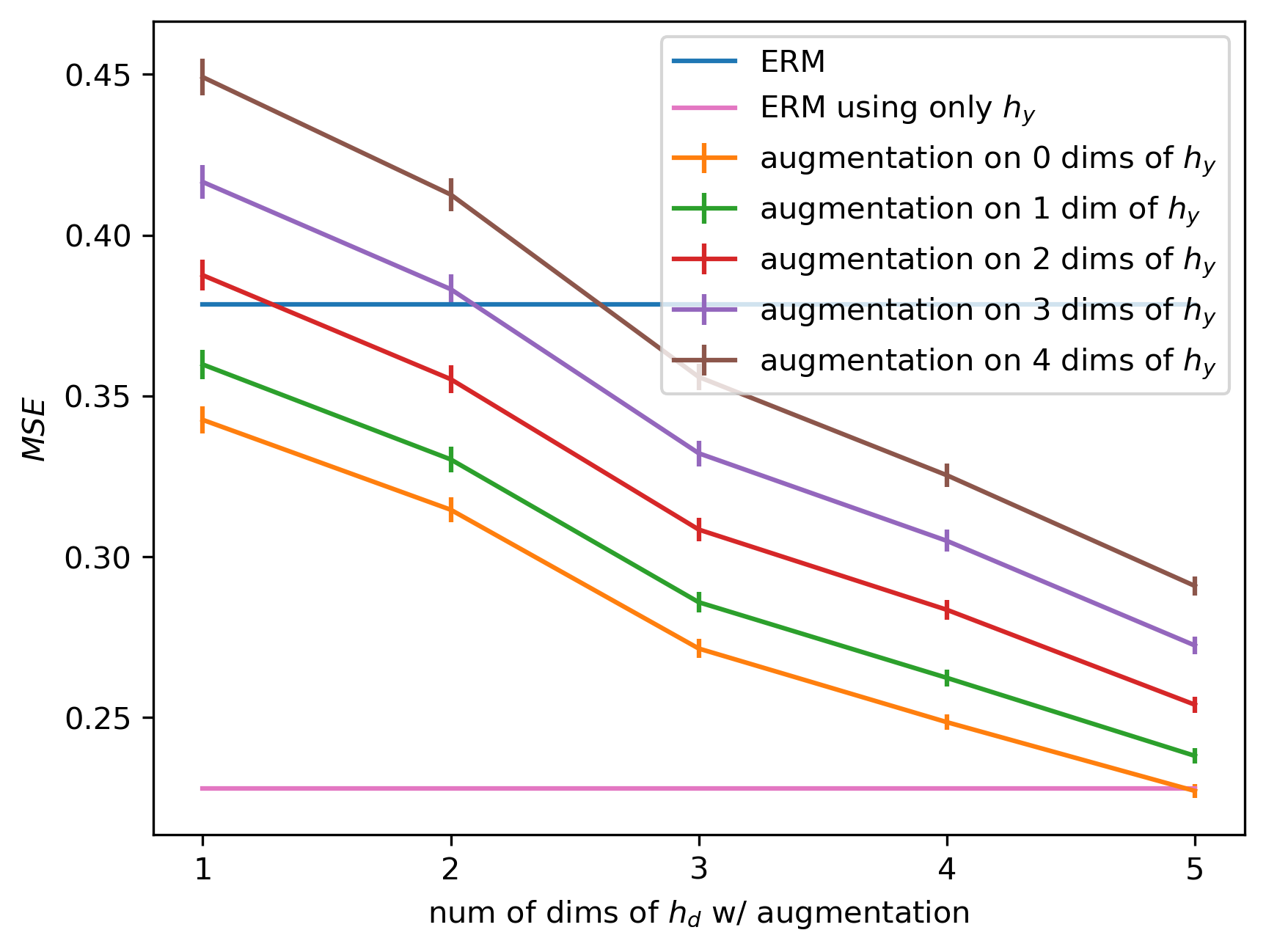}
    \caption{Results on synthetic data.}
    \label{fig:toy_data_results}
\end{figure}

We choose $c$, $d$, $y$, $h_d$ and $h_y$ to be five dimensional vectors. Furthermore, we sample the elements of the square matrices $W_{c\xrightarrow{}d}$, $W_{c\xrightarrow{}y}$, $W_{d\xrightarrow{}h_d}$ and $W_{y\xrightarrow{}h_y}$ from $\mathcal{N}(0, I)$. In all of our experiments $\sigma_c = I$ and $\sigma = 0.1 \cdot I$. The task is to regress $\sum_i^5 y_i$ from $x$, where $x=[h_d, h_y]$, a 10 dimensional feature vector.
During training the data is generated using the DAG in Figure \ref{fig:toy_dag_and_scm} (left), where due the confounder $c$ the features $h_d$ and $y$ are spuriously correlated. During testing we set $d := \mathcal{N}(0, I)$, keeping $W_{c\xrightarrow{}d}$, $W_{c\xrightarrow{}y}$, $W_{d\xrightarrow{}h_d}$ and $W_{y\xrightarrow{}h_y}$ the same as during training. As a result, features $h_d$ and $y$ are no longer correlated. A model relying on features $h_d$ will not be able to generalize well to the test data.
In all experiments, we use linear regression to minimize the empirical risk. We choose to add noise sampled from a uniform distribution $U[-10, 10]$ as our data augmentation technique. We vary the number of dimensions of $h_d$ as well as of $h_y$ that are augmented. Each experiment is repeated 50 times, in Figure \ref{fig:toy_data_results} we plot the mean of the mean square error (MSE) together with the standard error.

In Figure \ref{fig:toy_data_results}, we see that ERM using only features $h_y$ (pink line) achieves the lowest MSE. Next, we apply data augmentation to one, two, three, four, and five dimensions of $h_d$ while keeping $h_y$ unchanged (orange line). We find that if data augmentation is applied to all five dimensions of $h_d$ we can match the MSE of ERM with only features $h_y$. In this case, we are satisfying the condition in Equation \ref{eq:data_aug}. Furthermore, we find that unsurprisingly the MSE of models trained with data augmentation applied to features $h_y$ increases (green, red, purple, and brown line). However, we can see that as long as we apply data augmentation to at least three dimensions of $h_d$ the resulting MSE is lower than ERM using all features $h_d$ and $h_y$ (blue line).
Perhaps the most surprising result of this experiment is that there exist conditions under which applying data augmentation to features caused by $d$ \textit{and} features caused by $y$ will result in better generalization performance compared to ERM using all features.

\subsection{Rotated MNIST}
\label{sec:rotated_mnist}
We construct the rotated MNIST dataset following \citet{ferrari_deep_2018}. 
This dataset consists of four different domains $d$ and ten different classes $y$, each domain corresponds to a different rotation angle: $d = \{ 0^\circ, 30^\circ, 60^\circ, 90^\circ\}$. We first randomly select a subset of images $x$ from the MNIST training dataset and afterward apply the rotation to each image of the subset. For the next domain, we randomly select a new subset. To guarantee the
variance of $p(y)$ among the domains, the number of training examples for each digit class $y$ is randomly chosen from a uniform distribution 
$U[80, 160]$.

For each experiment three of the domains are selected for training and one domain is selected for testing. For the test domain, the corresponding rotation is applied to the 10000 examples of the MNIST testset. In Table \ref{tab:rot_mnist}, we compare data augmentation in combination with ERM to ERM, a Domain Adversarial Neural Network (DANN) \citep{ganin_domain-adversarial_2016} and a Conditional Domain Adversarial Neural Network (CDANN) \citep{ferrari_deep_2018}. All methods use a LeNet \citep{lecun_gradient-based_1998} type architecture and we repeat each experiment 10 times. First, we use SDA to find the best data augmentation technique, where we use the same LeNet model and training procedure for the domain classifier and only samples from the training domains. The data augmentation with the lowest domain accuracy in all four cases, where we leave out one of the domains for testing, is 'rotation'. In addition, we perform an ablation study showing that SDA reliably picks the most suitable hyperparameters, the results can be found in Table \ref{tab:ablation_rot_mnist} in the Appendix. Second, we apply random rotations between $0^\circ$ and $359^\circ$ to the images $x$ during training, denoted by DA. If we assume $h_d$ to be equal to the rotation angle of the MNIST digit in a given image $x$, applying random rotations to $x$ is equal to a noise intervention on $h_d$, see Equation \ref{eq:data_aug}. As described in the Method section, applying random rotations to $x$ can be understood as randomly sampling elements $g$ from the two-dimensional rotation group $SO(2)$. Note that elements $g \in SO(2)$ act trivially on $h_y$: Rotations do not change the digit shapes. The result is a training dataset where $d$ and $y$ are independent. In Table \ref{tab:rot_mnist}, we see that the results of DA are similar for all four test domains. Furthermore, we find that DA outperforms ERM, DANN, and CDANN, where CDANN is specially designed for the case where $d$ and $y$ are spuriously correlated.

\begin{table}[h]
    \caption{Results on Rotated MNIST results. Average accuracy for ten seeds.}
    \centering
\begin{tabular}{ c|c c c c }
Target &ERM & DANN & CDANN & SDA\\
\hline
 $0^\circ$ &75.4 &77.1 &78.5 &\textbf{96.1}\\ 
 $30^\circ$ &93.4 &94.2 &94.9 &\textbf{95.9}\\ 
 $60^\circ$ &94.5 &95.2 &95.6 &\textbf{95.7}\\ 
 $90^\circ$ &79.6 &83.0 &84.0 &\textbf{95.9}\\
 \hline
 Ave &85.7 &87.4 &88.3 & \textbf{95.9}
\end{tabular}
    \label{tab:rot_mnist}
\end{table}

\label{sec:PACS}
\begin{table*}[h]
    \centering
    \caption{Results on PACS dataset. Average accuracy for five seeds.}
\begin{tabular}{ c|c c c c c c c c c c c }
Target &ERM & CDANN & L2G & GLCM & SSN & IRM & REx & MetaReg & JigSaw & SDA\\
\hline
 A &63.3 &62.7 &66.2  &66.8  &64.1 &67.1 &67.0  &69.8  & 67.6 & \textbf{70.45}\\ 
 C &63.1 &69.7 &66.9  &69.7  &66.8 &68.5 &68.0  &70.4  & \textbf{71.7} &  68.49\\ 
 P &87.7 &78.7 &88.0  &87.9  &90.2 &89.4 &89.7  &\textbf{91.1}  & 89.0 & 88.35\\ 
 S &54.1  &64.5 &59.0  &56.3  &60.1 &57.8 &59.8  &59.3  & 65.2 &  \textbf{72.24}\\
 \hline
 Ave &67.1 &68.9  &70.0 &70.2 &70.3 &70.7 &71.1  &72.6  & 73.4 & \textbf{74.9}
\end{tabular}
    \label{tab:pacs}
\end{table*}

\subsection{Colored MNIST}
\label{sec:colored_mnist}
Following \citet{arjovsky_invariant_2019}, we create a version of the MNIST dataset where the color of each digit is spuriously correlated with a binary label $y$. We construct two training domains and one test domain where the digits of the original MNIST classes '0' to '4' are labeled $y=0$ and the digits of the classes '5' to '9' are are labeled $y=1$. Subsequently, for $25\%$ of the digits we flip the label $y$. We now color digits which are labeled $y=0$ red and digits which are labeled $y=1$ green. Last, we flip the color of a digit with a probability of $0.2$ for the first training domain and with a probability of $0.1$ for the second training domain. In the case of the test domain, the color of a digit is flipped with a probability of $0.9$. By design, the original MNIST class of each digit ('0' to '9') is a direct cause of the new label $y$ whereas the color of each digit is a descendant of the new label $y$.

The DAG of the colored MNIST, shown in Appendix Figure \ref{fig:dag_color_mnist}, deviates slightly from the DAG in Figure \ref{fig:gen_w_conf}, nonetheless the reasoning in the Method section is still valid.
In Table \ref{tab:colored_MNIST}, we see that while ERM is performing well on the training domains it fails to generalize to the test domain since it is using the color information to predict $y$. In contrast, IRM \citep{arjovsky_invariant_2019} and REx \citep{krueger_out--distribution_2020} generalizes well to the test domain. Again, we use SDA to find the appropriate data augmentations. We use the same MLP and training procedure as in \citet{arjovsky_invariant_2019} for the domain classifier. We want to highlight that SDA only relies on samples from the two training domains whereas the hyperparameters of IRM and REx where fine-tuned on samples from the test domain as described in \citet{krueger_out--distribution_2020}. In case of the colored MNIST dataset the selected data augmentations are 'hue' and 'translate', denoted by DA. As described in the Method section, applying random permutations to the hue value of $x$ is equivalent to randomly sampling and applying elements $g$ from permutation group $SO(2)$. We argue that elements $g$ do not change $h_y$: high-level features that contain information about the shape of each digit.
In our experiment, we use the same network architecture and training procedure as described in \citet{arjovsky_invariant_2019}. Each experiment is repeated 10 times. We find that DA can successfully weaken the spurious confounding influence of the domain $d$ on $y$, see Table \ref{tab:colored_MNIST}. 
\begin{table}[h]
    \centering
    \small
    \caption{Results on Colored MNIST. Average accuracy $\pm$ standard deviation for ten seeds.}
\begin{tabular}{ c|c c c c}
Acc &ERM &IRM & REx &SDA\\
\hline
 Train &\textbf{87.4 $\pm$ 0.2} &70.8 $\pm$ 0.9 & 71.5 $\pm$ 1.0 &72.1 $\pm$ 0.4\\ 
 Test &17.1 $\pm$ 0.6 & 66.9 $\pm$ 2.5 &68.7 $\pm$ 0.9  &\textbf{74.1 $\pm$ 0.9}\\ 
\end{tabular}
    \label{tab:colored_MNIST}
\end{table}

\subsection{PACS}
The PACS dataset \citep{li_deeper_2017} was introduced as a strong benchmark dataset for domain generalization methods that features large domain shifts. The dataset consists of four domains: $d$ = ['photo' (P), 'art-painting' (A), 'cartoon' (C), 'sketch' (S)], i.e., each image style is viewed as a domain. The numbers of images in each domain
are 1670, 2048, 2344, 3929 respectively. There are seven classes: $y$ = [dog,
elephant, giraffe, guitar, horse, house, person]. We fine-tune an AlexNet-model \citep{krizhevsky_imagenet_2012}, that was pre-trained on ImageNet, using ERM in combination with data augmentation. We apply SDA to select the data augmentation for the following experiment. For the domain classifier we fine-tune an AlexNet-model as described above. In addition, we use a cross-validation procedure where we leave one domain out and use the three domains for training. SDA determines four data augmentation techniques to be usefull: 'brightness', 'contrast', 'saturation', and 'hue'. In combination these four augmentations are commonly called color jitter or color perturbations. By randomly applying color perturbations we are weakening the spurious confounding influence of $h_d$ on $y$, as described in the Method section. In Table \ref{tab:pacs}, we compare DA to various domain generalization methods: CDANN \citep{ferrari_deep_2018}, L2G \citep{li_learning_2017}, GLCM \citep{wang_learning_2018}, SSN \citep{mancini_best_2018}, IRM \citep{arjovsky_invariant_2019}, REx \citep{krueger_out--distribution_2020}, MetaReg \citep{balaji_metareg_2018}, JigSaw \citep{carlucci_domain_2019}, where all methods use the same pre-trained AlexNet-model. We repeat each experiment 5 times and report the average accuracy. We find that DA obtains the highest average accuracy. The biggest performance gains of DA compared to ERM are on the test domains ’art painting’ and ’sketch’. For example, the domain ’sketch’ consists of black sketches of the seven object classes on white background, see Figure \ref{fig:pacs}. Since the color of the object is not correlated with the class, a model relying on color features will generalize poorly to the ’sketch’ domain. However, by randomly changing the colors of the images in the training domains (’art painting’, ’cartoon’, ’photo’), we find that DA is able to generalize much better. 

\subsection{Ablation study: Using all data augmentation techniques}
We repeat the previous experiments on Rotated MNIST, Colored MNIST, and PACS using all data augmentation techniques listed in the Appendix. We compare the accuracy of a classifier trained using all data augmentation techniques to a classifier trained using SDA. In Table \ref{tab:all_da_vs}, we find that using all data augmentation techniques together results in a significant drop in performance for all three datasets: $25.4\%$ for Rotated MNIST, $8.7\%$ for Colored MNIST, and $16.1\%$ for PACS. We argue that there are some combinations of the randomly applied data augmentation techniques that will destroy features $h_y$ caused by the label $y$. Besides, we observe that there exist combinations of datasets and data augmentation techniques that lead to a drastic drop in performance on their own, e.g the PACS dataset and random rotations. We argue that a model trained without random rotations exploits the fact that, e.g, the orientation of an animal or person is usually upright. This example shows that we cannot simply describe data augmentation as 'label-preserving transformations' since a rotated animal or person will still have the same label. We argue that we need to adopt a causal point of view to sufficiently describe the effect of data augmentation.

\begin{table}[h]
    \centering
    \small
    \caption{Comparison of a classifier trained using all data augmentation techniques and SDA. Average accuracy for all domains of each dataset. For details see Table \ref{tab:rot_mnist}, \ref{tab:colored_MNIST} and \ref{tab:pacs}.}
\begin{tabular}{ c|c c c c}
Dataset &All DA & SDA\\
\hline
 Rotated MNIST &70.5 &\textbf{95.9}\\ 
 Colored MNIST &65.4 &\textbf{74.1}\\ 
 PACS &58.8 &\textbf{74.9}\\ 
\end{tabular}
    \label{tab:all_da_vs}
\end{table}

\section{Conclusion}
In this paper, we present a causal perspective on the effectiveness of data augmentation in the context of domain generalization. By using an SCM we address a core problem of domain generalization: the spurious correlation of the domain variable $d$ and the target variable $y$. While in theory, we could intervene on the domain variable $d$, this solution is impractical since we assume that we only have access to observational data. However, we show that data augmentation can serve as a surrogate tool for simulating interventions on the domain variable $d$ and its children. 
Hereby, prior knowledge can be used to design data augmentation techniques that only act on the non-descendants of the target variable $y$. Furthermore, we show that randomly applying data augmentation can be understood as randomly sampling elements from common symmetry groups. In addition, we propose a simple algorithm to select suitable augmentation techniques from a given list of transformations. We use a domain classifier to measure how well each augmentation is able to weaken the causal link between the domain $d$ and $h_d$ high-level features caused by $d$. We evaluated this approach on four different datasets and were able to show that empirical risk minimization in combination with accurately designed data augmentation results in good generalization performance. The analysis in this paper could be further used to design data augmentation to simulate interventional datasets for domain generalization methods by exploiting intervention-augmentation equivariance.

\section{Acknowledgments}
The authors want to thank Leon Bottou and Ishaan Gulrajani for their help with IRM. In addition we want to thank David Lopez-Paz, Jörn-Henrik Jacobsen, Ben Glocker, Nick Pawlowski, and Daniel Castro for useful discussions.

Maximilian Ilse was funded by the Nederlandse Organisatie
voor Wetenschappelijk Onderzoek (Grant DLMedIa: Deep
Learning for Medical Image Analysis).
\newpage
\bibliography{aaai21}

\newpage
\section{Appendix}
\subsection{Additional details for SDA}
All data augmentations are implemented using the \texttt{TORCHVISION.TRANSFORMS} module of PyTorch \citep{paszke_pytorch_2019}. We choose the range of the hyperparameters of the augmentations in such a way that they do not destroy all information in $x$, e.g., setting the brightness of all pixels to 0 or translating all pixels by the full image width. In all experiments we use the following data augmentations:
\begin{itemize}
\tiny
    \item \texttt{'brightness':\\ torchvision.transforms.ColorJitter(brightness=1.0, contrast=0, saturation=0, hue=0)}
    \item \texttt{'contrast':\\ torchvision.transforms.ColorJitter(brightness=0, contrast=10.0, saturation=0, hue=0)}
    \item \texttt{'saturation':\\ torchvision.transforms.ColorJitter(brightness=0, contrast=0, saturation=10.0, hue=0)}
    \item \texttt{'hue':\\ torchvision.transforms.ColorJitter(brightness=0, contrast=0, saturation=0, hue=0.5)}
    \item \texttt{'rotation':\\ torchvision.transforms.RandomAffine([0, 359], translate=None, scale=None, shear=None, resample=PIL.Image.BILINEAR, fillcolor=0)}
    \item \texttt{'translate':\\ torchvision.transforms.RandomAffine(0, translate=[0.2, 0.2], scale=None, shear=None, resample=PIL.Image.BILINEAR, fillcolor=0)}
    \item \texttt{'scale':\\ torchvision.transforms.RandomAffine(0, translate=None, scale=[0.8, 1.2], shear=None, resample=PIL.Image.BILINEAR, fillcolor=0)}
    \item \texttt{'shear':\\ torchvision.transforms.RandomAffine(0, translate=None, scale=None, shear=[-10., 10., -10., 10.], resample=PIL.Image.BILINEAR, fillcolor=0)}
    \item \texttt{'vflip':\\ torchvision.transforms.RandomVerticalFlip(p=0.5)}
    \item \texttt{'hflip':\\ torchvision.transforms.RandomHorizontalFlip(p=0.5)}
\end{itemize}

\subsubsection{Ablation study on rotated MNIST}
We will demonstrate now that SDA can also be used to find the most suitable hyperparameters for the data augmentations used in this paper. In this example we focus on the rotated MNIST dataset and the data augmentation 'rotate'. We use the same experimental setup as described in the rotated MNIST experiment. We choose $\{30^\circ, 60^\circ, 90^\circ\}$ as the training domains and $0^\circ$ as the test domain. We compare five sets of hyperparameters, where each set defines the range from which the rotation angle is uniformily sampled. In Table \ref{tab:ablation_rot_mnist}, we find that the hyperparameters $[0^\circ, 359^\circ]$ lead to the lowest domain accuracy, i.e., simulate an intervention on $h_d$ the best.

\begin{table}[h]
    \centering
    \caption{Comparing domain accuracy on rotated MNIST for five different sets of the data augmentation 'rotate'. Average $\pm$ standard error over five seeds.}
\begin{tabular}{ c|c}
Hyperparameter &domain accuracy\\
\hline
 $[-15^\circ, 15^\circ]$ & 92.60 $\pm$ 0.98\\
 $[-45^\circ, 45^\circ]$ & 82.63 $\pm$ 0.89\\
 $[-90^\circ, 90^\circ]$ & 69.79 $\pm$ 0.91\\
 $[0^\circ, 180^\circ]$ & 63.16 $\pm$ 1.51\\
 $[0^\circ, 359^\circ]$ & 51.70 $\pm$ 2.21\\
\end{tabular}
    \label{tab:ablation_rot_mnist}
\end{table}

\subsubsection{Results of domain classifier on each dataset}
For each dataset, we train a domain classifier using the same architecture and training procedure as used for the label classifier. We only use samples from the training domains and repeat each experiment five times. In Table \ref{tab:domain_acc_results}, we show the domain accuracy for each of the datasets. In the case of rotated MNIST, we perform four experiments where each of the domains $d = \{ 0^\circ, 30^\circ, 60^\circ, 90^\circ\}$ is used for testing once, while the remaining three domains are used for training. For each individual experiment SDA returns the augmentation 'rotate' as the most suitable. In Table \ref{tab:domain_acc_results}, we show the average of the four experiments that where each repeated five times. In the case of colored MNIST, the training and test domains are fixed therefore we only conducted one experiment. We show the average of the one experiment that was repeated five times. For PACS, we perform four experiments where each of the domains $d = \{$'photo', 'art painting', 'cartoon', 'sketch'$\}$ is used for testing once, while the remaining three domains are used for training. We use cross validation over all four experiments to select the data augmentation. In Table \ref{tab:domain_acc_results}, we show the average of the four experiments that where each repeated five times.
\begin{table*}[h]
    \centering
    \caption{Domain accuracy for each dataset. Average $\pm$ standard error.}
\begin{tabular}{ c|c c c}
Data Augmentation & rotated MNIST & Colored MNIST & PACS \\
\hline
 'brightness'      & 98.45 $\pm$ 0.24  & 50.1524 $\pm$ 0.1527  & 96.46 $\pm$ 0.37\\
 'contrast'        & 98.64 $\pm$ 0.23  & 50.1470 $\pm$ 0.0506  & 96.41 $\pm$ 0.37\\
 'saturation'      & 98.95 $\pm$ 0.21  & 50.1894 $\pm$ 0.0593  & 96.03 $\pm$ 0.43\\
 'hue'             & 98.66 $\pm$ 0.36  & 50.0006 $\pm$ 0.0028  & 96.32 $\pm$ 0.41\\
 'rotation'        & 64.70 $\pm$ 2.21  & 50.0024 $\pm$ 0.0030  & 96.59 $\pm$ 0.39\\
 'translation'     & 90.84 $\pm$ 1.65  & 50.0004 $\pm$ 0.0008  & 96.82 $\pm$ 0.34\\
 'scale'           & 91.42 $\pm$ 1.34  & 50.2082 $\pm$ 0.1327  & 97.00 $\pm$ 0.29\\
 'shear'           & 91.48 $\pm$ 1.14  & 50.2252 $\pm$ 0.1531  & 96.82 $\pm$ 0.34\\
 'vertical flip'   & 88.79 $\pm$ 0.50  & 50.1560 $\pm$ 0.0140  & 96.88 $\pm$ 0.34\\
 'horizontal flip' & 91.98 $\pm$ 0.29  & 50.4060 $\pm$ 0.0274  & 96.54 $\pm$ 0.33\\
\end{tabular}
    \label{tab:domain_acc_results}
\end{table*}

\subsection{Colored MNIST}
\begin{figure}[H]
\begin{minipage}{.24\textwidth}
    \centering
    \resizebox{4cm}{4.1cm}{
    \input{tikz_plots/colored_mnist.tex}
    }
\end{minipage}
\begin{minipage}{.2\textwidth}
    \centering
    \resizebox{4cm}{4.1cm}{
    \input{tikz_plots/colored_mnist_intervention}
    }
\end{minipage}
\caption{Left: DAG of the data generating process for the colored MNIST dataset. Right: The same DAG after intervention on $h_d$. Interventional nodes are squared.}
\label{fig:dag_color_mnist}
\end{figure}
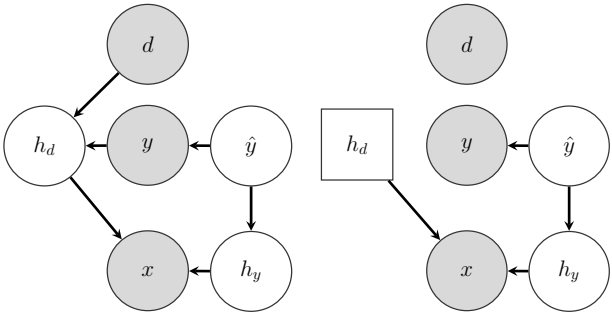

The DAG of the data generating process for the colored MNIST experiment is shown in Figure \ref{fig:dag_color_mnist} (left), where $d$ is the domain, $y$ is the binary label, $\hat{y}$ is the original MNIST label, $h_d$ are high-level color features caused by $d$ and $y$, $h_y$ are high-level shape features caused by $\hat{y}$, and $x$ is the observed image. In the case of the colored MNIST dataset the spurious correlation between $d$ and $y$ is the result of the collider $h_d$ (that itself is a parent of the observed node $x$). While the cause of the spurious correlation between $d$ and $y$ is different, the reasoning in the Method section is still valid. In Figure \ref{fig:dag_color_mnist} (right), we show that in theory an intervention on $h_d$ will remove the spurious correlation between $d$ and $y$. We argue that an intervention on $h_d$ can be simulated by data augmentation, we present experimental evidence in the Experiment section.

\subsection{PACS}
\begin{figure}[htb]
    \centering
    \includegraphics[scale=0.2]{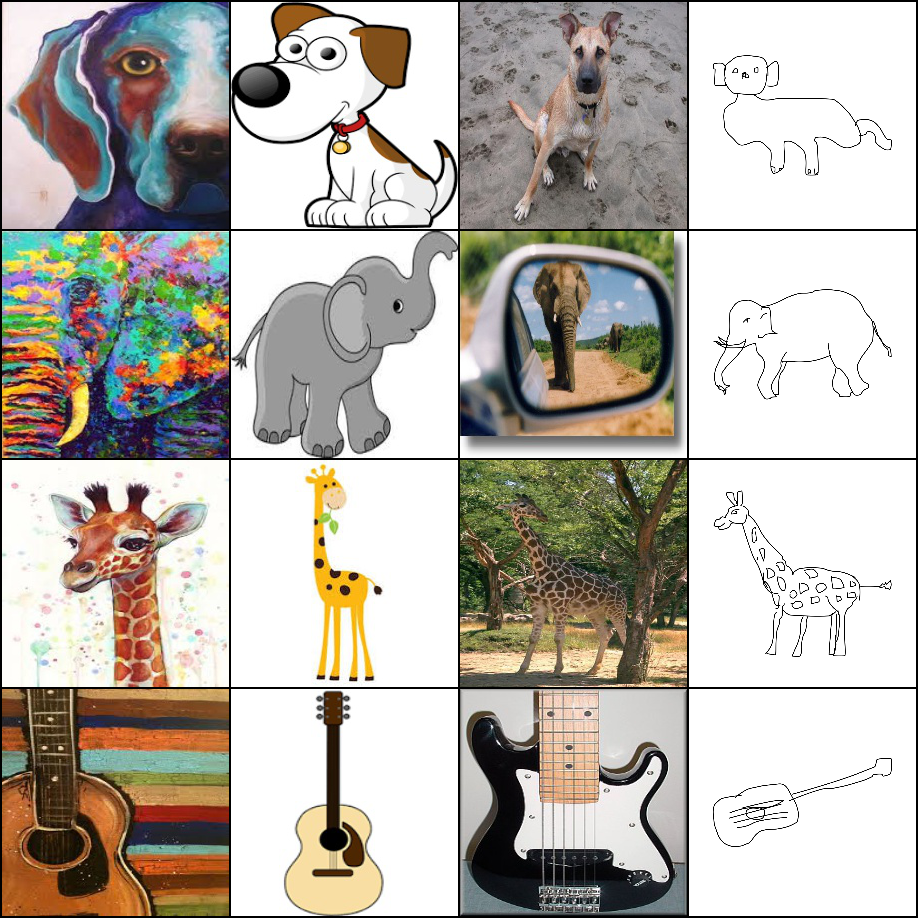}
    \caption{Samples from the first four classes ('dog', 'elephant', 'giraffe', 'guitar') for each domain (art-painting (A), cartoon (C), photo (P), sketch (S)) of the PACS dataset \citep{li_deeper_2017}.}
    \label{fig:pacs}
\end{figure}

Example images of the PACS dataset, see Figure \ref{fig:pacs}

\subsection{Linear example of intervention-augmentation equivariance}
\label{sec:linear_example}
A simple linear example can be constructed where the domain $d$ causes a specific ordering in $h_d$ that is spuriously correlated with the label $y$. In addition, $G$ is the permutation group and $g \in G$ acts as a permutation matrix $A$ on $x$, i.e.,  $Ax = g \cdot x$. In particular, we assume that $f_X(\cdot)$ is a linear transformation
\begin{align}
x = f_X(h_d, h_y) = C h_d + D h_y + e,
\label{eq:causal}
\end{align}
where $x, h_d, h_y, e$ are vectors and $C, D$ are matrices correspondingly sized. The data augmentation can be expressed as a linear transformation of the form
\begin{align}
    x_\text{aug} = \text{aug}_A(x) = A x,
\label{eq:da}
\end{align}
where $A$ is a correspondingly sized matrix sampled from the set of all permutation matrices. Combining Equation \ref{eq:causal} and \ref{eq:da}, we obtain
\begin{align}
    x_\text{aug} &= A x \nonumber\\
    &= AC h_d + AD h_y + Ae \nonumber\\
    &= C(C^{-1}AC h_d) + AD h_y + Ae \nonumber\\
    &= f_X(\doit_A(h_d), h_y).
    \label{eq:linear}
\end{align}
We find that if that $AD = D$ and $Ae = e$, i.e., $D$ and $e$ are permutation invariant, the transformation $Ax = g \cdot x$ successfully simulates the noise intervention $\doit_A(h_d) := C^{-1}AC h_d$ (with slight abuse of notation), i.e., we find that it satisfy the intervention-augmentation equivariance condition.

\subsection{Causality}
\label{sec:intro_causality}
What follows is a brief introduction of causal concepts that are used throughout this paper. It hopefully makes the paper more self-contained, as well as more accessible for readers that encounter these concepts for the first time. For an in-depth introduction please see \citet{pearl_causal_2009} or \citet{Peters2017}.

\subsubsection{Structural causal models}
\begin{figure}
    \centering
    \resizebox{8cm}{7.5cm}{\input{tikz_plots/dag_ccc.tex}}
    \caption{Top to bottom: chain, confounder, collider, chain with intervention on $y$.}
    \label{fig:dag_ccc}
\end{figure}
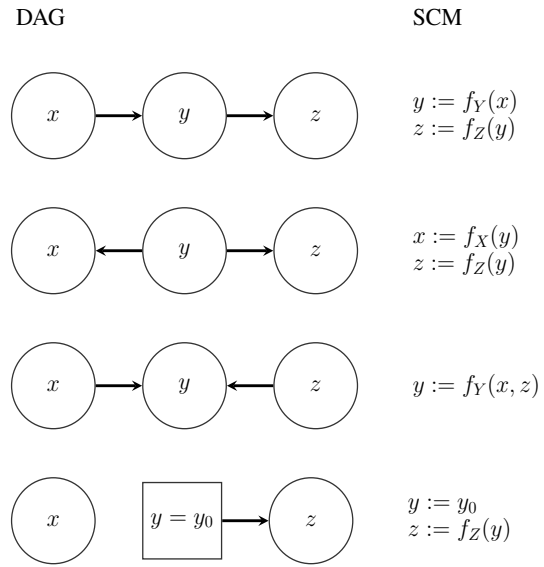

We say that a set of variables $x_1,\dots,x_l$ causes a variable $y$ if \emph{intervening} on any of the $x_m$ changes the distribution of $y$.
This is usually different from (conditional) \emph{observational} dependence between the $x_m$ and $y$.
Structural Causal Models (SCMs) are used to formalize those causal interactions between variables. 
We need to distinguish between two types of variables: exogenous and endogenous variables. Exogenous variables can be seen as an entry point to our SCM (and are usually unobserved independent random variables). 
The endogenous variables $x_m$ are then determined by the causal mechanisms, which are formalized via functional relations:  $x_m=f_m(x_{\text{pa}_m})$, where $x_{\text{pa}_m}$ is the tuple of the so-called parent variables of $x_m$. 
These relations of an SCM induces a corresponding graphical model. In this paper, we only deal with acyclic relationships, leading to Directed Acyclic Graphs (DAGs) as part of a Bayesian network. In Figure \ref{fig:dag_ccc}, we see three SCMs and their corresponding DAGs. Note that the direction of the arrows indicates the causal direction.

The SCMs in Figure \ref{fig:dag_ccc} are considered to be the three main building blocks of every causal model: chain, confounder, and collider. Where each of them introduces a different (conditional in-)dependence structure. First row: In case of a chain the variables $x$ and $z$ become conditionally independent if we condition on the center variable $y$, i.e., $p(z|x,y)=p(z|y)$. 
Second row: An observed confounder $y$ can introduce spurious correlation between its two children variables $x$ and $z$, i.e., we may have $p(x,z)\neq p(x)p(z)$. If we condition on the confounding variable $y$ they become conditionally independent again, i.e., $p(z|x,y)=p(z|y)$ and $p(x|z,y)=p(x|y)$. 
Third row: In case of an unobserved collider $y$ the two parent variables are independent, $p(x,z) = p(x)p(z)$. However, if we condition on $y$ they may become conditionally dependent, i.e., $p(x,z|y) \neq p(x|y)p(z|y)$.

\subsubsection{Interventions}
In its simplest form an intervention can be described as setting a variable $y$ to a constant value, e.g., $y=y_0$ irrespective of its parent variables. The result of such an intervention on the SCM of a chain and the corresponding DAG can be seen in the bottom row of Figure \ref{fig:dag_ccc}. In this example, the variable $y$ becomes independent of its parent variable $x$, i.e., we are replacing the function assignment $y=f(x)$ with $y=y_0$, effectively deleting the function $f(\cdot)$ and the corresponding arrow in the DAG. Using the $do$-operator \citep{pearl_causal_2009} we can write the resulting interventional distribution as follows: $p(z|x,\doit(y=y_0))=p(z|\doit(y=y_0))$.
In this paper, we use a special form of interventions, so-called noise or stochastic interventions \citep{peters_causal_2016}. Instead of setting the intervened variable to a fixed value, we randomize the values of $y$, i.e., $\doit(y=\xi)$, where $\xi$ is sampled from a noise distribution $N_\xi$.

\subsection{Domain generalization}
\subsubsection{Domain generalization via invariant feature representation}
\label{sec:learning_invariant_representations}

Arguably, the most commonly used approach in domain generalization relies on learning domain invariant features. The learning of invariant features can be achieved by mapping an input $x$ to intermediate features $z$ that are uninformative of the domain $d$, i.e., $p(z|d=i) = p(z|d=j)$. At the same time, the intermediate features $z$ are optimized for a low prediction error on all training domains. This results in finding a saddle point for the setting commonly referred to as domain adversarial learning \citep{ganin_domain-adversarial_2016}. It is assumed that such $z$ will generalize well to the test domain and, thus, result in a low test error.

Recent work of \citet{zhao_learning_2019}, \citet{johansson_support_2019} and \citet{arjovsky_invariant_2019}, in the context of domain adaptation, shows that enforcing $p(z|d=i) = p(z|d=j)$ is not necessarily leading to a low test error if the domains $d$ and targets $y$ are spuriously correlated, i.e., $p(y|d=i)$ $\neq$ $p(y|d=j)$. We now extend the findings of \citet{zhao_learning_2019} to domain generalization. 

As shown in \citet{zhao_learning_2019} an information-theoretic lower bound can be derived for the domain adaptation case. The bound "demonstrates that learning invariant representations could lead to a feature space where the joint error on both domains is large."
We provide a straightforward extension of this bound for the domain generalization case.

Introduction of notation:
\begin{itemize}
    \item $\rvx$: input 
    \item $\rvz$: intermediate representation
    \item $\hat{y}$: output 
    \item function composition: $x \xrightarrow[]{g} z \xrightarrow[]{h} \hat{y}$
    \item $y$: true label
    \item $h$: function mapping $x$ to $z$
    \item $g$: function mapping $z$ to $\hat{y}$
    \item $\text{JSD}$: Jensen-Shannon divergence
    \item $\epsilon^{d=i}$: empirical risk on domain $d=i$
\end{itemize}

Besides, we need the following two lemmas from \citet{zhao_learning_2019}. Proofs can be found in \citet{zhao_learning_2019}.

\textbf{Lemma 4.6:}
\begin{align}
    \text{JSD}(p(\hat{y}|d=i) || p(\hat{y}|d=j))\\ \leq \text{JSD}(p(z|d=i) || p(z|d=j)),
\end{align}
where $p(\hat{y}|d=i)$ are the marginal distributions of the output in domain $d=i$ and $p(z|d=i)$ are the marginal distributions of the intermediate representation in domain $d=i$.

\textbf{Lemma 4.7:}
\begin{align}
    \text{JSD}(p(y|d=i) || p(\hat{y}|d=i)) \leq \sqrt{\epsilon_i(h\circ g)},
\end{align}
i.e., how well is my output distribution matching the true labels distribution.\\

We start with the pairwise sum of Jensen-Shannon divergence between all $N$ training domains and the $N+1$ test domain
\begin{align}
    \sum_{1\leq i < j \leq N+1}\text{JSD}(p(y|d=i) || p(y|d=j)).
\end{align}

Since $\text{JSD}$ is a metric we can write
\begin{align}
    \sum_{1\leq i < j \leq N+1} \text{JSD}(p(y|d=i)|| p(y|d=j))\\
    \leq \sum_{1\leq i < j \leq N+1} \text{JSD}(p(\hat{y}|d=i)|| p(\hat{y}|d=j))\\
    + 2\sum_k^{N+1} \text{JSD}(p(y|d=k)|| p(\hat{y}|d=k)).
\end{align}

 Using Lemma 4.6 we get
 \begin{align}
    \sum_{1\leq i < j \leq N+1} \text{JSD}(p(y|d=i)|| p(y|d=j))\\
    \leq \sum_{1\leq i < j \leq N+1} \text{JSD}(p(z|d=i)|| p(z|d=j))\\
    + 2\sum_k^{N+1} \text{JSD}(p(y|d=k)|| p(\hat{y}|d=k)).
\end{align}

 Using Lemma 4.7 we get
  \begin{align}
    \sum_{1\leq i < j \leq N+1} \text{JSD}(p(y|d=i)|| p(y|d=j))\\
    \leq \sum_{1\leq i < j \leq N+1} \text{JSD}(p(z|d=i)|| p(z|d=j))\\
    + 2\sum_k^{N+1} \sqrt{\epsilon^{d=k}(h\circ g)}.
\end{align}

Extracting terms that belong to the test domain $d = N + 1$ leads to
\begin{align}
    \sum_{l=1}^N \text{JSD}(p(y|d=l)|| p(y|d=N+1))\\
    +\sum_{1\leq i < j \leq N} \text{JSD}(p(y|d=i)|| p(y|d=j)) \\
    \leq \sum_{l=1}^N \text{JSD}(p(z|d=l)|| p(z|d=N+1)) \\
    +\sum_{1\leq i < j \leq N} \text{JSD}(p(z|d=i)|| p(z|d=j))\\
    + 2\sqrt{\epsilon^{d=N+1}(h\circ g)}
    + 2\sum_k^{N} \sqrt{\epsilon^{d=k}(h\circ g)}
\end{align}

Assuming we find a perfect intermediate representation $z$ for all $N$ training domains and the test domain $d = N+1$ (assuming such an $z$ exists) we are left with
\begin{align}
    \sum_{l=1}^N \text{JSD}(p(y|d=l)|| p(y|d=N+1))\\
    +\sum_{1\leq i < j \leq N} \text{JSD}(p(y|d=i)|| p(y|d=j)) \\
    \leq 2\sqrt{\epsilon^{d=N+1}(h\circ g)}
    + 2\sum_k^{N} \sqrt{\epsilon^{d=k}(h\circ g)}
    \label{eq:bound}
\end{align}
We see that, as it was the case for domain adaptation, that the joint risk across all domains (training and test) is lower bounded by the pairwise divergence of the marginal label distribution of all domains. Given the existence of an unobserved confounder as seen in Figure 1 the marginal label distribution are unlikely to match.

However, there exists a multitude of domain generalization methods that do not explicitly address the problem of hidden confounders \citep{balaji_metareg_2018, carlucci_agnostic_2018, carlucci_domain_2019, ding_deep_2018, ghifary_domain_2015, ilse_diva_2019, li_learning_2017, mancini_best_2018, motiian_unified_2017, shankar_generalizing_2018, tzeng_deep_2014, wang_learning_2018}. However, the majority of these methods are evaluate on benchmark datasets, e.g., VLCS \citep{hutchison_undoing_2012} or PACS \citep{li_deeper_2017}, where the domain $d$ and the target $y$ are confounded. As shown in Equation \ref{eq:bound}, this can result in poor generalization performance. Nonetheless, we cannot rule out the possibility that some of these methods are implicitly able to deal with confounders, thus achieving good generalization performance.

To the best of our knowledge, there are currently very few methods that address the issue of spuriously correlated domains $d$ and targets $y$ \citep{arjovsky_invariant_2019, heinze-deml_conditional_2019, ferrari_deep_2018, krueger_out--distribution_2020}, where \citet{ferrari_deep_2018} extends the idea of domain adversarial learning to enforce conditional domain invariance, i.e., $p(z|y, d=i) = p(z|y, d=j)$.

\subsection{Data augmentation}
\label{sec:data_augmentation}
We will briefly summarize how data augmentation is currently viewed in the computer vision community, for a in-depth survey see \citet{shorten2019survey}. In computer vision data augmentation is seen as an effective technique for improving the performance on a variety of tasks such as image classification, object detection, and image segmentation.
In the image domain, data augmentation techniques can be roughly divided into two categories:
\begin{enumerate}
    \item Augmenting the geometry of an image: Commonly used transformations are rotations, horizontal and vertical flips, scaling, cropping, occlusion, and elastic deformations.
    \item Augmenting the color of an image: Random values are added or subtracted from the color channels of an image. Instead of applying this transformation directly in the RGB colorspace, other color spaces like CIELAB and HSL are commonly used \citep{tellez_quantifying_2019}.
\end{enumerate}
Data augmentation is a combination of the transformation listed above that are randomly applied to all images during training.

\subsubsection{Data augmentation in application-focused research areas}
\label{sec:histo}
\label{sec:robots}
In the following, we give a summary of two examples of the successful application of data augmentation for domain generalization in medical imaging and robotics. We want to highlight that in both examples the actual task and the domains are spuriously correlated.

\paragraph{Histopathology}
The high variability of the appearance of histopathology images is a major obstacle for the deployment of automatic image analysis systems. The variability of appearance is the result of a multitude of preparation steps that are applied to the specimen: cutting, fixating, staining, and scanning. Each step introduces its own artifacts. This leads to different color distributions among histopathology laboratories. \citet{tellez_quantifying_2019} perform a detailed comparison of commonly used data augmentation, see Appendix Figure \ref{fig:histo}. The augmentation techniques consist of random rotation and flipping, random color perturbation, and color normalization. These augmentation techniques are compared on a dataset composed of histopathology images from nine different laboratories. We argue that there exists a hidden confounder that spuriously correlates the staining and scanner artifacts (caused by the laboratories) and the abnormalities in the tissue (caused by the diseases). By augmenting the color of the histopathology images \citet{tellez_quantifying_2019} are able to learn features that are invariant to the laboratories. Furthermore, \citet{tellez_quantifying_2019} find that random color perturbation outperforms color normalization. We argue that random color perturbation simulates noise interventions, whereas color normalization tries to simulate interventions where the color of a histopathology image is set to a fixed value. As described in the Method section, this requires to first estimate the color distribution of the original histopathology image which is a challenging problem. As a result, data augmentation in the form of random color perturbation is better suited to simulate interventional data.
\begin{figure}[h]
    \centering
    \includegraphics[scale=0.5]{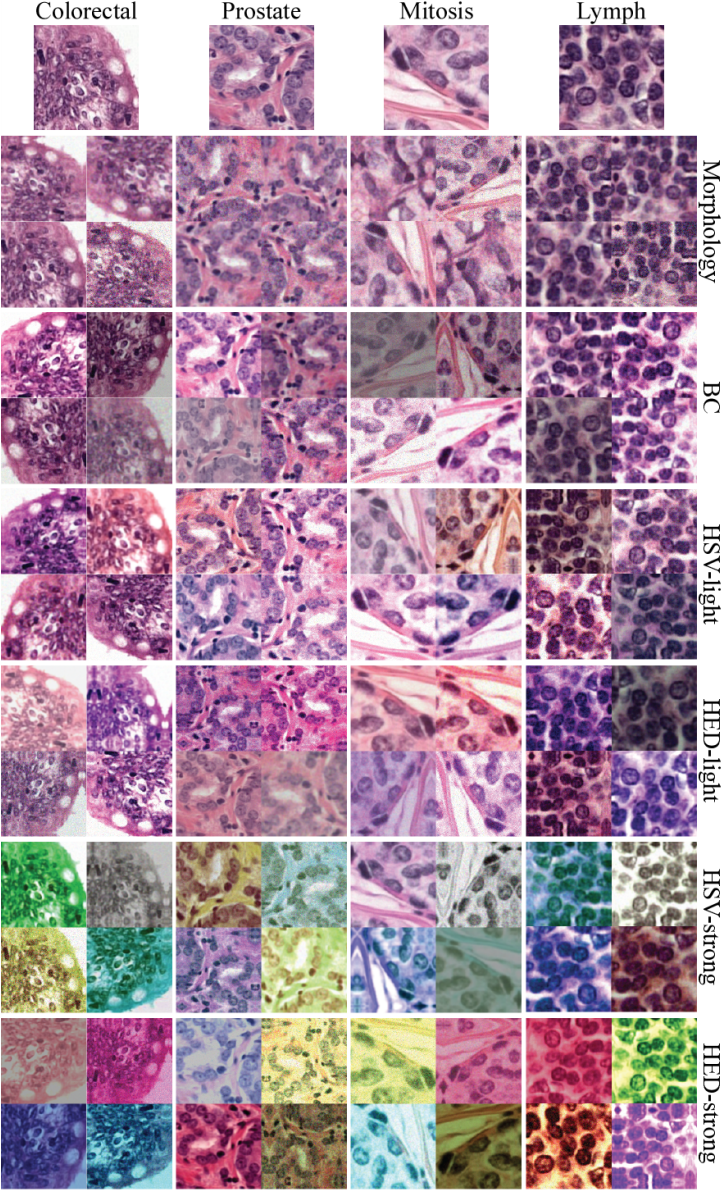}
    \caption{Domain randomization histopathology, taken from \citet{tellez_quantifying_2019}}
    \label{fig:histo}
\end{figure}

\paragraph{Robotics}
Performing robotic learning on physical hardware is often not feasible due to: (\textit{i}) the large number of training samples that are required, and (\textit{ii}) potential damage to the hardware if the learning relies on random exploration. Therefore, learning in a physics simulator is of great interest. While learning in a simulator is cheap and safe, we are facing a new problem, namely, how to overcome the so-called \textit{reality gap}, i.e., the differences between simulation and the real world. In \citet{tobin_domain_2017} they focus on a robotic manipulation task that involves a robotic arm and eight 3D objects that are placed on a table. In this scenario, a neural network is used to detect the location of an object.  To be able to generalize from the simulation to the real world, \citet{tobin_domain_2017} apply a variety of data augmentation techniques to the simulator, e.g., randomization of position and texture of all objects on the table, textures of the table, floor, skybox, and robot,
and the addition of random noise. 
We argue that there exists a hidden confounder that introduces a spurious correlation between, e.g., the lighting conditions and the location of the objects on the table. By applying heavy data augmentation during the training process they are able to generalize to unseen lighting conditions in the real world.
\begin{figure}[h]
    \centering
    \includegraphics[scale=0.2]{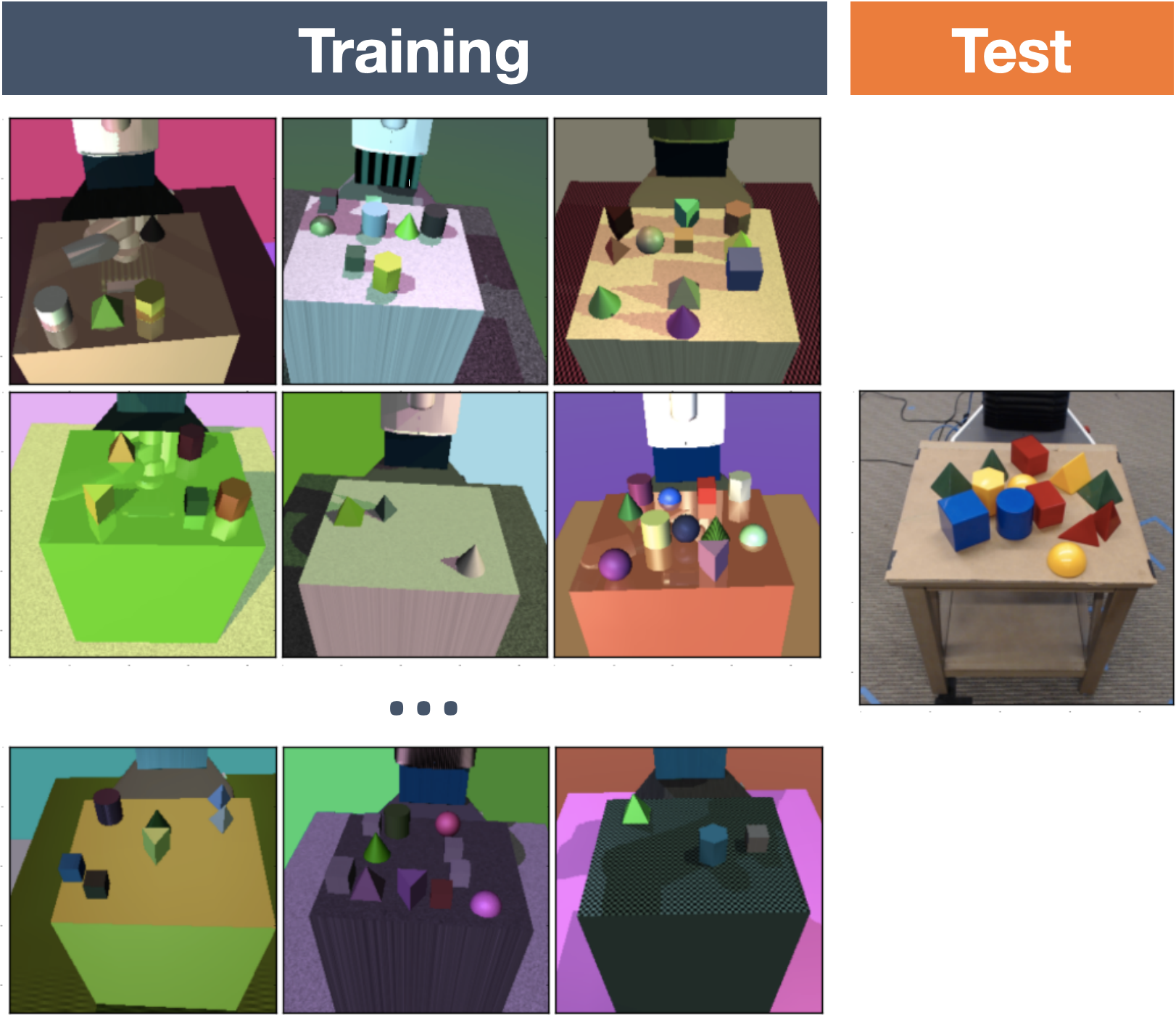}
    \caption{Domain randomization in robotics, taken from \citet{tobin_domain_2017}}
    \label{fig:robots}
\end{figure}

\end{document}

%% file: math_commands.tex
\usepackage{amsmath,amsfonts,bm}

\def\eqref#1{equation~\ref{#1}}

\def\1{\bm{1}}

\def\rvx{{\mathbf{x}}}

\def\rvz{{\mathbf{z}}}

\DeclareMathAlphabet{\mathsfit}{\encodingdefault}{\sfdefault}{m}{sl}
\SetMathAlphabet{\mathsfit}{bold}{\encodingdefault}{\sfdefault}{bx}{n}

\newcommand{\doit}{\mathrm{do}}
\newcommand{\aug}{\mathrm{aug}}

%% file: tikz_plots/confounder_proxies.tex
\begin{tikzpicture}
\LARGE
\node [node] (c) {$c$};
\node [nodeobserved, below left = of c, yshift=1cm](d) {$d$};
\node [node, below = of d](h_d) {$h_d$};
\node [nodeobserved, below right = of c, yshift=1cm](y) {$y$};
\node [node, below = of y](h_y) {$h_y$};
\node [nodeobserved, below right = of h_d, yshift=1cm](x) {$x$};

\draw [arrow] (y) -- (h_y);
\draw [arrow] (h_y) -- (x);

\draw [arrow] (d) -- (h_d);
\draw [arrow] (h_d) -- (x);

\draw [arrow] (c) -- (d);
\draw [arrow] (c) -- (y);
\end{tikzpicture}

%% file: tikz_plots/intervention_on_d.tex
\begin{tikzpicture}
\LARGE
\node [node] (c) {$c$};
\node [nodeinter, below left = of c, yshift=1.1cm](d) {$d$};
\node [node, below = of d](h_d) {$h_d$};
\node [nodeobserved, below right = of c, yshift=1cm](y) {$y$};
\node [node, below = of y](h_y) {$h_y$};
\node [nodeobserved, below right = of h_d, yshift=1cm](x) {$x$};

\draw [arrow] (y) -- (h_y);
\draw [arrow] (h_y) -- (x);

\draw [arrow] (d) -- (h_d);
\draw [arrow] (h_d) -- (x);

\draw [arrow] (c) -- (y);
\end{tikzpicture}

%% file: tikz_plots/intervention_on_hd.tex
\begin{tikzpicture}
\LARGE
\node [node] (c) {$c$};
\node [nodeobserved, below left = of c, yshift=1.1cm](d) {$d$};
\node [nodeinterwhite, below = of d](h_d) {$h_d$};
\node [nodeobserved, below right = of c, yshift=1cm](y) {$y$};
\node [node, below = of y](h_y) {$h_y$};
\node [nodeobserved, below right = of h_d, yshift=1cm](x) {$x$};

\draw [arrow] (y) -- (h_y);
\draw [arrow] (h_y) -- (x);

\draw [arrow] (h_d) -- (x);

\draw [arrow] (c) -- (y);
\draw [arrow] (c) -- (d);
\end{tikzpicture}

%% file: tikz_plots/data_augmentation.tex
\begin{tikzpicture}
\LARGE
\node [node] (c) {$c$};
\node [nodeobserved, below left = of c, yshift=1cm](d) {$d$};
\node [node, below = of d](h_d) {$h_d$};
\node [nodeobserved, below right = of c, yshift=1cm](y) {$y$};
\node [node, below = of y](h_y) {$h_y$};
\node [nodeobserved, below right = of h_d, yshift=1cm](x) {$x$};
\node [nodeobserved, below = of x](x_aug) {$x_\text{aug}$};

\draw [arrow] (y) -- (h_y);
\draw [arrow] (h_y) -- (x);

\draw [arrow] (d) -- (h_d);
\draw [arrow] (h_d) -- (x);

\draw [arrow] (c) -- (d);
\draw [arrow] (c) -- (y);
\draw [arrow] (x) -- (x_aug);
\end{tikzpicture}

%% file: tikz_plots/commutative.tex
\begin{tikzpicture}
    
     

\Large
\node[fill=white](h1_h2) {$(h_d, h_y)$};
\node[fill=white, right = of h1_h2, xshift=1.1cm](x) {$x$};
\node[fill=white, below = of h1_h2, yshift=-1cm](h1tilde_h2) {$(\doit(h_d), h_y)$};
\node[fill=white, below = of x, yshift=-1.2cm](x_aug) {$x_\aug$};

\draw [arrow,|->] (h1_h2) -- node [above,midway] {$f_X$} (x);
\draw [arrow,|->] (x) -- node [above,midway,xshift=0.6cm,yshift=-0.45cm] {$\aug$} (x_aug);
\draw [arrow,|->] (h1_h2) -- node [above,midway,xshift=-0.5cm,yshift=-0.25cm] {$\doit$} (h1tilde_h2);
\draw [arrow,|->] (h1tilde_h2) -- node [above,midway,xshift=-0.15cm] {$f_X$} (x_aug);

\end{tikzpicture}



%% file: tikz_plots/experiment_dag.tex
\begin{tikzpicture}
\LARGE
\node [node] (c) {$c$};
\node [nodeobserved, below left = of c](d) {$d$};
\node [nodeobserved, below right = of c](y) {$y$};

\node [nodeobserved, below = of d](h1) {$h_d$};
\node [nodeobserved, below = of y](h2) {$h_y$};

\draw [arrow] (y) -- (h2);
\draw [arrow] (d) -- (h1);
\draw [arrow] (c) -- (d);
\draw [arrow] (c) -- (y);
\end{tikzpicture}

%% file: tikz_plots/colored_mnist.tex
\begin{tikzpicture}
\LARGE
\node [nodeobserved] (d) {$d$};
\node [node, below left = of d](hd) {$h_d$};
\node [nodeobserved, below = of d, yshift=0.55cm](y) {$y$};
\node [node, below right = of d](y_hat) {$\hat{y}$};
\node [node, below = of y_hat](hy) {$h_y$};
\node [nodeobserved, below = of y](x) {$x$};

\draw [arrow] (d) -- (hd);
\draw [arrow] (hd) -- (x);
\draw [arrow] (y_hat) -- (y);
\draw [arrow] (y) -- (hd);
\draw [arrow] (y_hat) -- (hy);
\draw [arrow] (hy) -- (x);
\end{tikzpicture}

%% file: tikz_plots/colored_mnist_intervention.tex
\begin{tikzpicture}
\LARGE
\node [nodeobserved] (d) {$d$};
\node [nodeinterwhite, below left = of d, yshift=0.2cm](hd) {$h_d$};
\node [nodeobserved, below = of d, yshift=0.55cm](y) {$y$};
\node [node, below right = of d](y_hat) {$\hat{y}$};
\node [node, below = of y_hat](hy) {$h_y$};
\node [nodeobserved, below = of y](x) {$x$};

\draw [arrow] (hd) -- (x);
\draw [arrow] (y_hat) -- (y);
\draw [arrow] (y_hat) -- (hy);
\draw [arrow] (hy) -- (x);
\end{tikzpicture}

%% file: tikz_plots/dag_ccc.tex
\begin{tikzpicture}
\LARGE
\node [node] (x) {$x$};
\node [node, right = of x](y) {$y$};
\node [node, right = of y](z) {$z$};
\draw [arrow] (x) -- (y);
\draw [arrow] (y) -- (z);

\node[fill=white,text width=4cm, right = of z](t) {$y:=f_Y(x)$\\ $z:=f_Z(y)$};

\node [node, below = of x](x1) {$x$};
\node [node, right = of x1](y1) {$y$};
\node [node, right = of y1](z1) {$z$};
\draw [arrow] (y1) -- (x1);
\draw [arrow] (y1) -- (z1);

\node[fill=white,text width=4cm, right = of z1] {$x:=f_X(y)$\\ $z:=f_Z(y)$};

\node [node, below = of x1](x2) {$x$};
\node [node, right = of x2](y2) {$y$};
\node [node, right = of y2](z2) {$z$};
\draw [arrow] (x2) -- (y2);
\draw [arrow] (z2) -- (y2);

\node[fill=white,text width=4cm, right = of z2] {$y:=f_Y(x,z)$};

\node [node, below = of x2] (x3) {$x$};
\node [nodeinterwhite, right = of x3](y3) {$y=y_0$};
\node [node, right = of y3](z3) {$z$};
\draw [arrow] (y3) -- (z3);

\node[fill=white,text width=4cm, right = of z3]() {$y:=y_0$\\ $z:=f_Z(y)$};

\node[fill=white,text width=4cm, above = of t] {SCM};
\node[fill=white,text width=1.6cm, above = of x, yshift=-0.2cm] {DAG};

\end{tikzpicture}